\documentclass{article}
\usepackage{iclr2027_conference,times}

\usepackage{amsmath,amsfonts,bm}

\def\eqref#1{equation~\ref{#1}}

\def\1{\bm{1}}

\DeclareMathAlphabet{\mathsfit}{\encodingdefault}{\sfdefault}{m}{sl}
\SetMathAlphabet{\mathsfit}{bold}{\encodingdefault}{\sfdefault}{bx}{n}

\usepackage{amsmath,amssymb,amsfonts,amsthm}
\usepackage{booktabs}
\usepackage{graphicx}
\usepackage{microtype}
\usepackage{multirow}
\usepackage{xcolor}
\usepackage{array}
\usepackage{placeins}
\usepackage{hyperref}
\usepackage{url}
\usepackage{etoc}
\renewcommand{\eqref}[1]{\textup{(\ref{#1})}}

\definecolor{jltorange}{RGB}{213,94,0}
\definecolor{jltblue}{RGB}{0,114,178}
\definecolor{jltgreen}{RGB}{35,139,69}
\hypersetup{
    colorlinks=true,
    linkcolor=jltblue,
    citecolor=jltblue,
    urlcolor=jltblue
}

\newcommand{\method}{JLT}
\newcommand{\FID}{FID-50K}
\newcommand{\IS}{IS}

\newcommand{\Lv}{\mathcal{L}_{v}}
\newcommand{\Lx}{\mathcal{L}_{x}}
\newtheorem{proposition}{Proposition}
\title{Equivalent Flows, Unequal Learning: Clean-Latent Prediction in Transformers}

\author{%
Funing Fu$^{1,*}$ \quad Tenghui Wang$^{2,*}$ \quad Guanyu Zhou$^{3}$ \quad Junyong Cen$^{1}$ \quad Qichao Zhu$^{4}$\\[0.30em]
{\small $^{1}$Independent Researcher \quad $^{2}$Wuhan University of Technology}\\[0.15em]
{\small $^{3}$Technology Innovation Institute \quad $^{4}$Hangzhou Jiyi Artificial Intelligence Co., Ltd.}\\[0.15em]
{\small\texttt{chinoll@chinoll.org} \quad \texttt{371062@whut.edu.cn}}\\
}
\date{}

\iclrfinalcopy
\begin{document}

\maketitle
\lhead{}
\renewcommand{\headrulewidth}{0pt}
\begingroup
\renewcommand{\thefootnote}{*}
\footnotetext[1]{Equal contribution.}
\endgroup

\vspace{-10pt}

\begin{figure}[htbp]
    \centering
    \includegraphics[width=\linewidth]{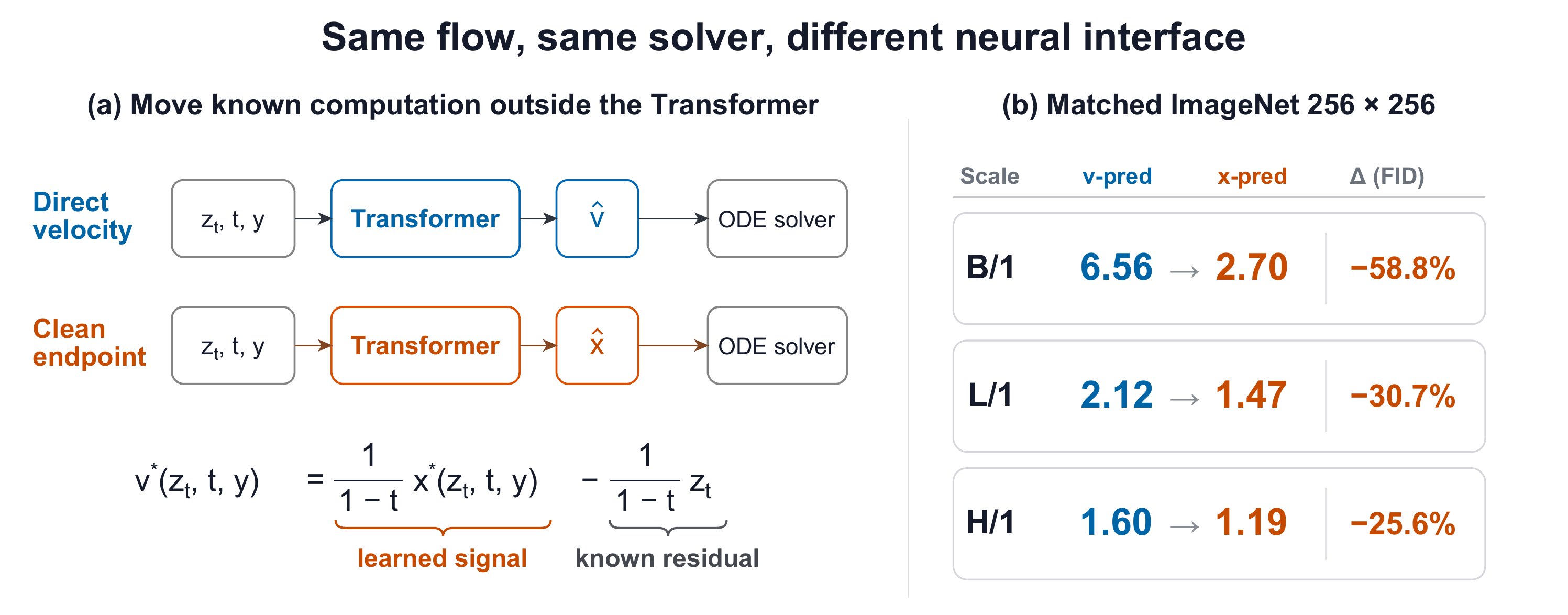}
    \caption{\textbf{The solver interface need not be the neural interface.}
    (a)~Direct velocity prediction learns the vector field internally, whereas clean-endpoint prediction exposes $\hat x$ directly to the Transformer and shifts the linear residual response $\hat v=(\hat x-z_t)/(1-t)$ to an exact readout, leaving the ODE solver interface identical.
    (b)~ImageNet $256\times256$ FID across B/1, L/1, and H/1 scales (Table~\ref{tab:matched}).}
    \label{fig:overview}
\end{figure}

\vspace{10pt}

\begin{abstract}
Flow samplers consume velocity, but the neural network can predict the clean endpoint and convert it to velocity through a fixed affine readout. We study this choice with \method{}, a latent Transformer in a frozen variational autoencoder (VAE) representation. For squared error, the optimal clean and velocity predictors are algebraically equivalent; a finite Transformer assigns different computation to its learned output under the two interfaces. A local Gaussian analysis identifies a known residual response supplied by the readout and isotropic target variance added by velocity prediction. Measured FLUX.2 channel spectra support this geometric distinction: 90\% of target variance occupies 83 of 128 clean directions versus 109 velocity directions. Under a matched velocity objective, clean prediction improves ImageNet FID-50K from 6.56 to 2.70 at Base scale and from 2.12 to 1.47 at Large scale, with lower FID at every measured Large checkpoint. Scaling clean prediction to 951M parameters reaches FID-50K 1.19 and IS 271.96. In addition, an objective ablation at Base scale shows that direct clean regression reaches FID-50K 2.38 without time-dependent error weighting. These results show how moving known computation outside the network changes learning under algebraically equivalent flow interfaces. Code: \url{https://github.com/akatsuki-neo/JLT}
\end{abstract}

\vspace{10pt}

\section{Introduction}
\label{sec:intro}

\begin{figure}[t]
    \centering
    \includegraphics[width=\linewidth]{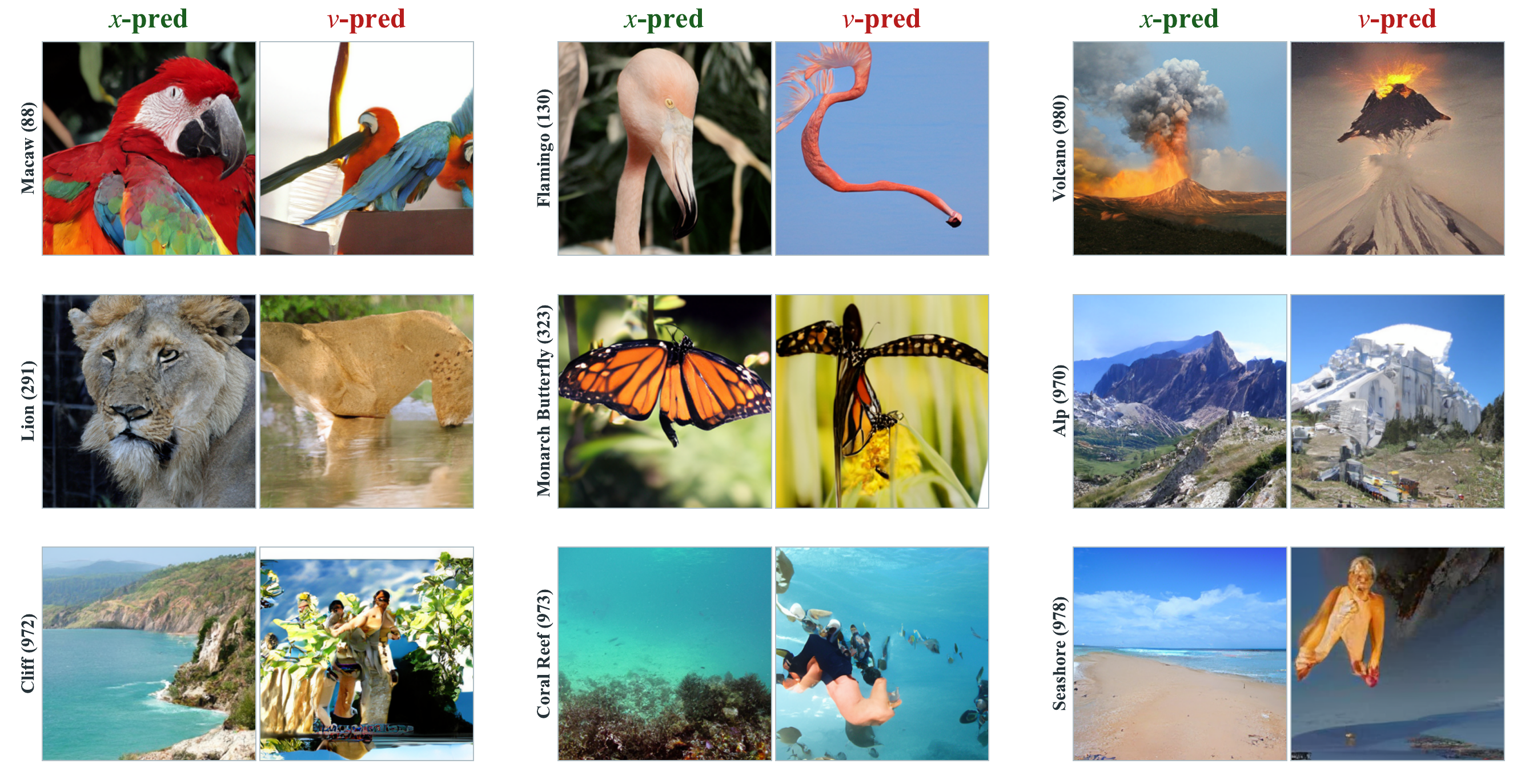}
    \caption{\textbf{Qualitative comparison of clean-$x$ and velocity-$v$ prediction under matched Base scale.}
    Randomly sampled class-conditional generations from matched B/1 models on ImageNet $256\times256$ with 50-step Heun sampling. The paired samples illustrate differences in global structure and sharpness.}
    \vspace{5pt}
    \label{fig:qualitative-main}
\end{figure}

An ordinary differential equation (ODE) solver needs a vector field. A neural network does not necessarily need to predict one.
Modern flow models often make these two interfaces identical: the solver consumes velocity, and the network is trained to predict velocity. For the linear paths $z_t=tx+(1-t)\epsilon$, however, a clean-endpoint prediction determines exactly the same velocity, $\hat v_x=\frac{\hat x-z_t}{1-t}$. At the level of the ideal flow, the two parameterizations are therefore equivalent. Our question is different: \textbf{does this equivalence survive the replacement of an ideal predictor by a finite neural network?}
For finite networks, algebraically equivalent outputs can induce different learned function classes. The affine conversion contains a known input residual and a time-dependent gain. A network that predicts $v$ must represent this response internally, whereas a network that predicts $x$ can delegate it to the analytic readout.
More generally, if a fixed transformation maps one valid output to another, applying that transformation outside a finite learned function need not produce the same parameterized family as learning the transformed quantity directly.
Given a corrupted state $z_t$ and time $t$, a prediction of the clean endpoint $x$ can be converted exactly into velocity by a fixed affine map. The generative flow can therefore remain unchanged while the quantity directly approximated by the neural network changes.

This distinction matters because algebraic equivalence is a statement about outputs, not about the finite parameterized functions used to produce them. For squared error, let $x^*(z_t,t,y)$ denote the optimal clean predictor. The corresponding optimal velocity predictor satisfies
\begin{equation}
    v^*(z_t,t,y)=\frac{x^*(z_t,t,y)-z_t}{1-t}.
    \label{eq:intro-decomposition}
\end{equation}
A direct-velocity Transformer approximates the entire right-hand side as one learned function. A clean-output Transformer can instead approximate $x^*$ and delegate the known input-dependent residual and time-dependent gain to the analytic readout. The solver receives the same quantity either way; what changes is the boundary between learned and known computation. Figure~\ref{fig:overview} summarizes this view.

Target geometry gives a second reason for the two neural interfaces to behave differently. Clean data in a learned representation can be strongly anisotropic. Velocity $v=x-\epsilon$ combines that structure with unit Gaussian noise, adding variation in every coordinate direction. In a local Gaussian model, the optimal clean predictor suppresses directions in which the clean latent has little variance, whereas the optimal velocity predictor retains a large negative residual response there. The analytic readout supplies this response exactly, leaving the Transformer to model the structured endpoint.

Clean prediction has recently proved effective in pixel-space denoising Transformers~\citep{li2025jit}, and related work studies generalized prediction targets~\citep{jin2026revisiting,gagneux2026training} and clean prediction in semantic representation spaces~\citep{zhang2026rit}. Our focus is the latent-flow setting in which generation takes place in a reconstruction-trained VAE representation~\citep{rombach2022high}. This setting is useful for isolating the parameterization question: the representation, Transformer architecture, and flow objective can be held fixed while the neural output interface changes.

\begin{figure}[t]
    \centering
    \includegraphics[width=\linewidth]{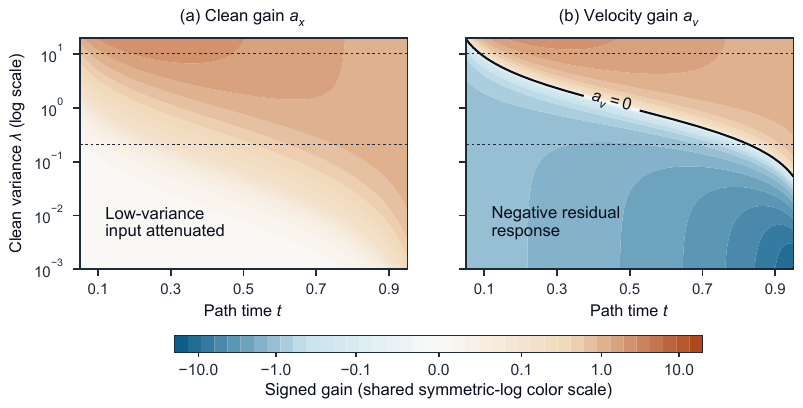}
    \caption{\textbf{The analytic readout supplies the residual response in low-variance directions.}
    Conditional-mean gains $a_x=t\lambda/[t^2\lambda+(1-t)^2]$ and $a_v=[t\lambda-(1-t)]/[t^2\lambda+(1-t)^2]$ for clean and velocity prediction, respectively, over path time $0<t<1$ and clean variance $\lambda$ (derived in Proposition~\ref{prop:geometry}). Both panels share a signed symmetric-log color scale. The solid curve marks zero velocity gain; horizontal dotted lines mark the measured FLUX channel-variance range, 0.21 to 10.50. As clean variance decreases, the clean predictor attenuates the input while the velocity response becomes increasingly dominated by the known residual term.}
    \label{fig:regression-gains}
\end{figure}

We instantiate this idea with \method{}, a latent Transformer that predicts the clean latent and converts it analytically to velocity.\footnote{Code: \url{https://github.com/akatsuki-neo/JLT}} On ImageNet $256\times256$ with a frozen FLUX.2 VAE, matched Base models improve \FID{} from 6.56 with direct velocity to 2.70 with clean output, and matched Large models improve from 2.12 to 1.47. Figure~\ref{fig:qualitative-main} shows paired random samples from the matched B/1 models. At Huge scale (951.3M parameters), clean prediction achieves \FID{} 1.19 and \IS{} 271.96. Across scales, the measured FLUX latent spectrum matches the geometric premise of the analysis: clean variation is concentrated, while adding the velocity noise floor spreads variance across substantially more channel directions.

Our contributions are threefold:
\begin{itemize}
    \item We separate the neural prediction interface from the flow solver interface. For linear flows, an exact clean-to-velocity readout moves known residual computation outside the Transformer while preserving the same vector field consumed by the sampler.
    \item We characterize how this reparameterization changes the learned function and target geometry. A local Gaussian analysis exposes the low-variance residual response, and measurements on frozen FLUX.2 latents show the predicted covariance shift and loss of spectral concentration.
    \item Under matched velocity objectives, clean prediction improves FID at Base and Large scales. At Huge scale, the clean-output model reaches \FID{} 1.19 and \IS{} 271.96.
\end{itemize}

\section{Related Work}
\label{sec:related}

\paragraph{Prediction targets and objectives.}
Diffusion models have long admitted multiple prediction parameterizations. Denoising diffusion probabilistic models popularized noise prediction~\citep{ho2020denoising,nichol2021improved}, while velocity parameterization was introduced in the context of progressive distillation~\citep{salimans2022progressive}. EDM emphasizes that parameterization, preconditioning, loss weighting, schedules, and sampling are distinct design axes~\citep{karras2022elucidating}; subsequent work studies time-dependent weighting~\citep{choi2022perception,hang2023efficient,kingma2023understanding}. Flow matching~\citep{lipman2023flow}, rectified flow~\citep{liu2023flow}, and stochastic interpolants~\citep{albergo2023building,albergo2023stochastic} formulate generation through vector fields. Recent analyses study the interaction between weighting and parameterization~\citep{gagneux2026training} and the dimensionality of different prediction targets~\citep{jin2026revisiting}. We fix the flow objective and ask which algebraically valid quantity should be exposed directly to the network.

\paragraph{Connection to preconditioning.}
EDM writes its denoiser as a known skip connection plus a scaled network output, $D_\theta(u;\sigma)=c_{\mathrm{skip}}(\sigma)u+c_{\mathrm{out}}(\sigma)F_\theta(\cdot)$~\citep{karras2022elucidating}. The unregularized \method{} readout has the same structural form: $\hat v=(1-t)^{-1}f^x_\theta-(1-t)^{-1}z_t$. Within this preconditioning structure, we study the exact target conversion for a linear flow, its placement across the neural/solver interface, and the comparison of direct clean and velocity outputs within a fixed Transformer family. We note that clean reconstruction is the classical objective of denoising autoencoders~\citep{vincent2008extracting,vincent2010stacked,bengio2013generalized,alain2014regularized}, while residual denoisers learn the corruption component~\citep{zhang2017beyond}. Natural images have low intrinsic dimensionality~\citep{pope2021intrinsic}, a property reflected in diffusion scores~\citep{stanczuk2024diffusion}. JiT~\citep{li2025jit} revisits clean prediction for pixel-space Transformers. Our analysis emphasizes a complementary view: clean prediction places an exact affine transformation outside the learned function, which changes both the effective function class and the geometry of the supervised target.

\paragraph{Latent and representation-space generation.}
Latent diffusion~\citep{rombach2022high} established generation in pretrained autoencoder spaces, and DiT~\citep{peebles2023dit}, SiT~\citep{ma2024sit}, Stable Diffusion 3~\citep{esser2024scaling}, U-ViT~\citep{bao2023all}, DiffiT~\citep{hatamizadeh2023diffit}, and PixArt~\citep{chen2023pixart} develop Transformer-based generators in latent spaces. Other work changes or augments the representation itself: REPA~\citep{yu2025repa} and SRA~2~\citep{wang2026vaerepa} use representation alignment, while representation autoencoders~\citep{zheng2025diffusion,singh2026improved} generate in semantic feature spaces. RiT~\citep{zhang2026rit} applies clean prediction directly in frozen DINOv2 features~\citep{oquab2024dinov2}. DPM-Solver++~\citep{lu2022dpmsolverpp} uses data prediction as a sampling parameterization. In contrast, we freeze a reconstruction-trained VAE and the Transformer family and study the quantity predicted directly during learning, while keeping the vector-field interface to the solver fixed.

\begin{table}[t]
    \centering
    \small
    \setlength{\tabcolsep}{5.0pt}
    \begin{tabular}{lllc c}
    \toprule
    Scale & Network predicts & Loss & \FID{} $\downarrow$ & \IS{} $\uparrow$ \\
    \midrule
    \multirow{2}{*}{B/1}
     & velocity $v$ & $\Lv$ & 6.56 & 132.12 \\
     & clean latent $x$ & $\Lv$ & \textbf{2.70} & \textbf{213.89} \\
    \midrule
    \multirow{2}{*}{L/1}
     & velocity $v$ & $\Lv$ & 2.12 & 236.21 \\
     & clean latent $x$ & $\Lv$ & \textbf{1.47} & \textbf{255.28} \\
    \midrule
    \multirow{2}{*}{H/1}
     & velocity $v$ & $\Lv$ & 1.60 & \textbf{327.41} \\
     & clean latent $x$ & $\Lv$ & \textbf{1.19} & 271.96 \\
    \bottomrule
    \end{tabular}
    \vspace{5pt}
    \caption{\textbf{Parameterization comparison across model scales.} Within each scale, the training setup and velocity objective are matched; only the neural prediction target changes.}
    \vspace{5pt}
    \label{tab:matched}
    \end{table}

\section{Same Flow, Different Learning Problems}
\label{sec:parameterization}
\label{sec:jlt}

\subsection{The Solver Interface Need Not Be the Neural Interface}

Let $x\in\mathbb{R}^{D}$ be a clean latent from a fixed encoder and $\epsilon\sim\mathcal{N}(0,I)$ be Gaussian noise. We use the linear path
\begin{equation}
    z_t=t x+(1-t)\epsilon, \qquad t\in[0,1],
    \label{eq:path}
\end{equation}
whose sample-wise velocity is $v=x-\epsilon$. A direct-velocity model learns
\begin{equation}
    f^{v}_{\theta}(z_t,t,y)\rightarrow v,
    \label{eq:v-param}
\end{equation}
whereas \method{} learns the clean endpoint
\begin{equation}
    f^{x}_{\theta}(z_t,t,y)\rightarrow x,
    \qquad
    \hat v^{x}_{\theta}=\frac{f^{x}_{\theta}(z_t,t,y)-z_t}{1-t}.
    \label{eq:x-param}
\end{equation}
The identities in this section use the unregularized readout for $0<t<1$. Both parameterizations expose a velocity interface to an ODE solver. Endpoint regularization in the implementation is discussed separately in Appendix~\ref{app:implementation}.

The distinction can be stated without a Gaussian assumption. Under squared error, the Bayes-optimal predictors are conditional means,
\begin{equation}
    x^*(z_t,t,y)=\mathbb{E}[x\mid z_t,t,y],
    \qquad
    v^*(z_t,t,y)=\mathbb{E}[v\mid z_t,t,y].
\end{equation}
Since $v=(x-z_t)/(1-t)$ sample-wise, conditioning gives the exact identity
\begin{equation}
    v^*(z_t,t,y)=\frac{x^*(z_t,t,y)-z_t}{1-t}.
    \label{eq:bayes-readout}
\end{equation}
The readout therefore does not approximate an additional unknown function. It implements a known input- and time-dependent transformation of the clean conditional mean.

This changes the effective parameterized family seen by the solver. If $\mathcal{F}=\{f_\theta\}$ denotes the Transformer function class, clean prediction induces
\begin{equation}
    \mathcal{V}_{x}=\left\{(z,t)\mapsto \frac{f(z,t)-z}{1-t}:f\in\mathcal{F}\right\},
    \label{eq:function-class}
\end{equation}
whereas direct velocity uses $\mathcal{V}_{v}=\mathcal{F}$. An unrestricted oracle can represent either form, but a fixed finite architecture need not be closed under the affine readout. In architectural terms, the clean parameterization supplies an explicit, time-dependent residual connection from the input $z_t$ to the solver output, rather than asking the Transformer to realize that response internally. Output parameterization therefore decides which computation is learned by the Transformer and which is supplied exactly by the model. Appendix~\ref{app:conversions} lists the corresponding conversions among $x$, $\epsilon$, and $v$.

\begin{figure}[t]
    \centering
    \includegraphics[width=0.99\linewidth]{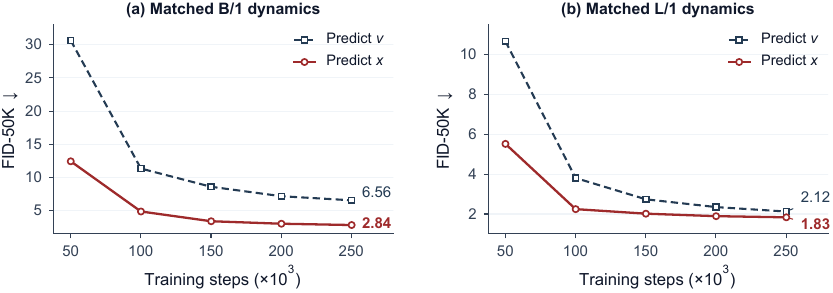}
    \caption{\textbf{Matched training dynamics across scales.} (a) Base scale (B/1, 130.5M params): clean-$x$ prediction maintains a 3.72 to 18.20 FID margin over direct velocity across all checkpoints; (b) Large scale (L/1, 458.1M params): clean-$x$ prediction is lower at every measured checkpoint, reaching 1.83 versus 2.12. All checkpoints within each pair are evaluated under the same setting.}
    \label{fig:training}
\end{figure}

\subsection{Target Geometry in an Anisotropic Latent Space}
\label{sec:geometry}

We next examine what each neural interface asks the network to regress. Consider a local Gaussian model $x\sim\mathcal{N}(0,\Sigma)$ with independent $\epsilon\sim\mathcal{N}(0,I)$. The eigenvalues of $\Sigma$ describe directions with strong or weak clean-latent variation.

\begin{proposition}[Target covariance and low-variance gain]
\label{prop:geometry}
For the path in Eq.~\eqref{eq:path} at $0<t<1$, the marginal target covariances are
\begin{equation}
    \operatorname{Cov}(x)=\Sigma,
    \qquad
    \operatorname{Cov}(v)=\Sigma+I.
    \label{eq:target-cov}
\end{equation}
For an eigen-direction of $\Sigma$ with eigenvalue $\lambda_i$ and $q_i=t^2\lambda_i+(1-t)^2$, the Bayes estimators are
\begin{align}
    \mathbb{E}[x_i\mid z_i]&=\frac{t\lambda_i}{q_i}z_i, \\
    \mathbb{E}[v_i\mid z_i]&=\frac{t\lambda_i-(1-t)}{q_i}z_i.
    \label{eq:bayes-estimators}
\end{align}
As $\lambda_i\rightarrow0$, the clean coefficient tends to $0$, while the velocity coefficient tends to $-1/(1-t)$.
\end{proposition}

The two limits reveal the division of labor created by the readout. Along a direction with little clean-latent variation, the optimal clean predictor attenuates the input. The optimal velocity predictor instead contains a large negative residual response. Equation~\eqref{eq:bayes-readout} shows where that response comes from: as the learned clean term vanishes, the fixed $-z_t/(1-t)$ part of the readout remains. The clean parameterization therefore exposes the structured endpoint to the Transformer and computes the residual response analytically.

Figure~\ref{fig:regression-gains} resolves this behavior across path time and clean variance. Writing the two coefficients as $a_x$ and $a_v$, the exact readout gives $a_v=(a_x-1)/(1-t)$. The clean coefficient is nonnegative, while the velocity coefficient changes sign at $\lambda=(1-t)/t$ and approaches $-1/(1-t)$ as the clean variance vanishes.

\paragraph{Spectral concentration.}
The covariance shift also changes how broadly the target variance is distributed. For descending eigenvalues $\lambda_i$ and $S=\operatorname{tr}(\Sigma)>0$, define $C_x(k)=\sum_{i=1}^{k}\lambda_i/S$. The fraction of velocity variance in its leading $k$ directions is
\begin{equation}
    C_v(k)=\alpha C_x(k)+(1-\alpha)\frac{k}{D},
    \qquad \alpha=\frac{S}{S+D}.
    \label{eq:spectral-mixture}
\end{equation}
Thus the velocity spectrum mixes the clean spectrum with a uniform spectrum. Since $C_x(k)\geq k/D$, we have $C_v(k)\leq C_x(k)$: adding isotropic noise makes the target variance less concentrated across the clean eigen-directions. This is a statement about target structure rather than numerical conditioning; $\Sigma+I$ can be better conditioned while requiring more directions to capture the same fraction of target variation.

\begin{figure}[t]
    \centering
    \includegraphics[width=\linewidth]{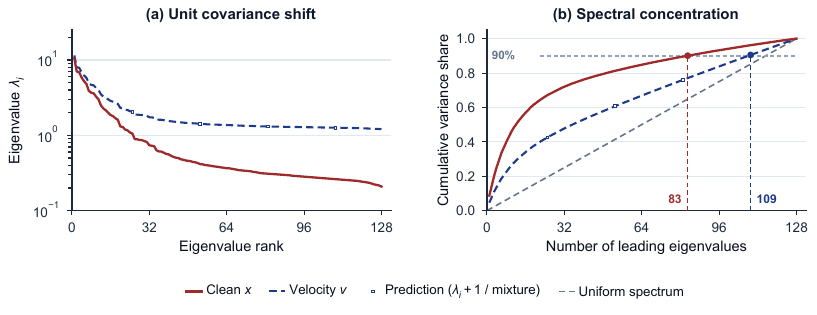}
    \caption{\textbf{Velocity spreads target variation across the frozen FLUX latent space.}
    All 128 channel eigenvalues are estimated over the ImageNet training set. (a) Measured clean and velocity spectra; open squares show the predicted shift $\lambda_i+1$. (b) Cumulative variance shares. Open squares show the mixture in Eq.~\eqref{eq:spectral-mixture}, and the diagonal is the uniform spectrum. Capturing 90\% of target variance requires 83 clean directions and 109 velocity directions. Prediction markers are spaced for readability; all directions enter the calculation.}
    \label{fig:latent-spectrum}
\end{figure}

\subsection{A Clean Neural Interface}

\method{} uses a standard latent Transformer. Images are encoded once by a frozen VAE, the latent grid is patchified, and time and class embeddings condition the Transformer blocks through adaptive modulation. The resulting sequence is processed by standard Transformer blocks and unpatchified into $\hat x_\theta$. The solver receives only $\hat v^x_\theta$ from Eq.~\eqref{eq:x-param}. The matched direct-velocity model uses the same input, Transformer architecture, conditioning, and solver; only the direct output quantity changes.

We use Base, Large, and Huge configurations following the plain JiT family~\citep{li2025jit}: depths 12, 24, and 32; hidden dimensions 768, 1024, and 1280; and 12, 16, and 16 attention heads. The /1 suffix denotes one latent-grid cell per token. The trainable parameter counts are 130.5M, 458.1M, and 951.3M.

\subsection{Training the Same Clean Output with Two Objectives}
\label{sec:losses}

Our primary comparison uses velocity-space regression. Before endpoint regularization, the clean-output objective is exactly velocity-space MSE for $0<t<1$. The clean-output network produces $\hat x_\theta$, the fixed readout produces $\hat v^x_\theta$, and the loss is
\begin{align}
    \Lv
    &=\mathbb{E}\left[\left\|\frac{\hat{x}_\theta-z_t}{1-t}-(x-\epsilon)\right\|_2^2\right] \\
    &=\mathbb{E}\left[\frac{\|\hat{x}_\theta-x\|_2^2}{(1-t)^2}\right].
    \label{eq:velocity-loss}
\end{align}
The second line shows the clean-error weighting induced by the unregularized velocity readout. The conditional residual variances obey the same scaling (Appendix~\ref{app:proof}). We also train the same clean-output model directly in endpoint space,
\begin{equation}
    \Lx=\mathbb{E}\left[\|\hat{x}_\theta-x\|_2^2\right].
    \label{eq:clean-loss}
\end{equation}
This provides a second training interface for the same neural output and the same sampler. \method{} denotes the clean-output parameterization, not a particular loss weighting: $\Lv$ and $\Lx$ train the same direct output and expose the same velocity interface at sampling time. In practice, the denominator in Eq.~\eqref{eq:velocity-loss} is clipped by $t_\epsilon=0.05$, and training times follow $\operatorname{logit}(t)\sim\mathcal{N}(-0.8,0.8^2)$. Appendix~\ref{app:implementation} gives the complete setup.

\begin{figure}[t]
    \centering
    \includegraphics[width=\linewidth]{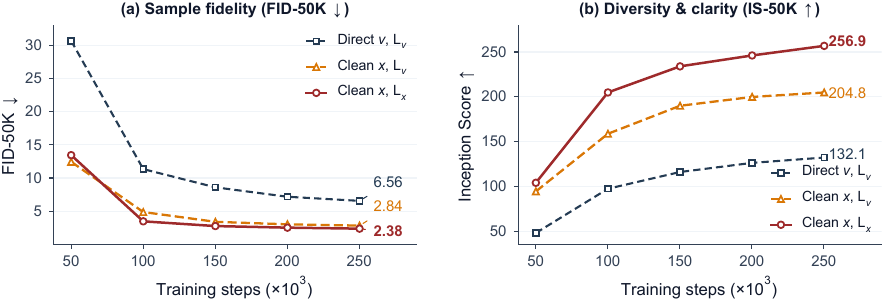}
    \caption{\textbf{Loss weighting ablation across training at Base scale.} (a) FID-50K trajectories for direct velocity under $\Lv$ and clean output under $\Lv$ or unweighted $\Lx$; final FIDs are 6.56, 2.84, and 2.38, respectively. (b) Corresponding IS-50K trajectories; final scores are 132.12, 204.83, and 256.88.}
    \label{fig:loss-weighting}
\end{figure}

\section{Experiments}
\label{sec:experiments}

We organize the experiments along four axes: matched output parameterization, latent covariance, loss weighting with a fixed clean output, and model scale. The matched comparison isolates the neural output while holding the flow objective fixed; the covariance measurement tests the geometric regime identified above; the loss ablation varies weighting without changing the direct output; and the cross-scale comparison across Base, Large, and Huge shows how the parameterization gap evolves with capacity. Together these experiments ask three questions: does changing only the neural output alter learning; does the frozen latent space exhibit the predicted target geometry; and does the clean interface remain effective without time-dependent error weighting?

\subsection{Setup}

We evaluate class-conditional ImageNet-1K generation at $256\times256$~\citep{deng2009imagenet,russakovsky2015imagenet}. Models use deterministic codes from a frozen FLUX.2 VAE~\citep{blackforestlabs2026flux2small}, train for 200 epochs (approximately 250K optimizer steps), and use class-drop probability 0.1. Evaluation generates 50K samples with 50-step Heun integration. We report Fréchet Inception Distance on 50K samples (\FID{}) and Inception Score (\IS{})~\citep{salimans2016is,heusel2017fid}. Appendix~\ref{app:implementation} gives the evaluation settings.

\subsection{Equivalent Flow, Unequal Learning}
\label{sec:matched}

Table~\ref{tab:matched} compares clean and velocity outputs. Within each scale, the models share the frozen encoder, Transformer, time distribution, optimizer, training budget, and velocity objective; their direct neural output differs.

At Base scale, directly predicting the clean latent improves \FID{} from 6.56 to 2.70 (a 3.86 point, or 58.8\%, reduction) and raises \IS{} from 132.12 to 213.89. At Large scale, where direct velocity already achieves \FID{} 2.12, clean output reaches 1.47 (a 0.65 point, or 30.7\%, reduction) and raises \IS{} from 236.21 to 255.28. At Huge scale (951.3M parameters), clean output attains \FID{} 1.19 and \IS{} 271.96, compared with \FID{} 1.60 for direct velocity.

The distinction is visible throughout optimization. Figure~\ref{fig:training} shows the matched B/1 and L/1 trajectories: at Base scale, clean output starts at \FID{} 12.44 versus 30.64 at 50K steps and converges to 2.84 versus 6.56; at Large scale, clean output starts at \FID{} 5.52 versus 10.67 and remains lower at every measured checkpoint, ending at 1.83 versus 2.12. On Qwen-VAE, matched B/1 models show the same FID ordering at the measured checkpoints, reaching \FID{} 6.54 versus 6.85 at 250K steps (Appendix~\ref{app:qwen-vae}, Figure~\ref{fig:qwen-training}; \citealp{wu2025qwenimagetechnicalreport}).

\subsection{Scale Dependence of the Parameterization Gap}
\label{sec:scale-dependence}

Table~\ref{tab:matched} compares Base, Large, and Huge models trained with the velocity objective $\Lv$. For B/1 and L/1, clean output reduces FID by 3.86 (58.8\%) and 0.65 (30.7\%), respectively. At H/1, clean output reaches 1.19, compared with 1.60 for direct velocity. The B/1 to L/1 gap narrows as model size increases; the H/1 result shows that clean prediction remains effective at 951.3M parameters. Table~\ref{tab:comparison} situates the 1.19 result alongside contemporary literature.

\subsection{Latent Covariance Spectrum: Does the Predicted Regime Hold?}
\label{sec:empirical-geometry}

We measure the covariance spectrum of the same frozen FLUX representation used for generation. All 1,281,176 ImageNet training images are encoded into $128\times16\times16$ latent tensors. Treating each spatial location as an observation, let $X\in\mathbb{R}^{N\times128}$ contain mean-centered latents after the same frozen channel normalization used in training, and let $E$ contain independent standard Gaussian noise. We estimate
\begin{equation}
    \widehat{\Sigma}_x=\frac{1}{N-1}X^\top X,
    \qquad
    \widehat{\Sigma}_v=\frac{1}{N-1}(X-E)^\top(X-E).
    \label{eq:empirical-covariance}
\end{equation}

Figure~\ref{fig:latent-spectrum} shows the predicted shift directly. The clean channel spectrum ranges from 10.50 to 0.21, a 50.36$\times$ span, whereas the velocity spectrum ranges from 11.50 to 1.21, closely following $\lambda_i+1$. Numerical conditioning improves after the shift: the covariance condition number falls from 50.36 to 9.52. Target concentration moves in the opposite direction, however: the effective rank rises from 35.10 to 78.97 out of 128 channels. The clean representation therefore remains strongly anisotropic after VAE compression, while velocity fills its low-variance tail with unit-scale variation and distributes variance more broadly.

The cumulative spectrum makes the same change directly visible. With clean trace 120.20 over 128 channels, Eq.~\eqref{eq:spectral-mixture} predicts that the velocity spectrum is a 48.4\%/51.6\% mixture of the clean and uniform spectra. The measured curve follows this prediction: 90\% of clean target variance is captured by 83 leading directions, while velocity requires 109. Thus the key distinction is not numerical conditioning ($\Sigma+I$ is better conditioned), but broader distribution of target variation across directions weakly supported by clean latents.

\subsection{Loss Weighting with a Fixed Clean Output}
\label{sec:weighting}

Output parameterization and loss weighting are distinct design choices. As shown in Eq.~\eqref{eq:velocity-loss}, training clean prediction under $\Lv$ induces a clean-error weighting of $(1-t)^{-2}$ before endpoint clipping. We additionally train the clean-output B/1 model with unweighted clean MSE ($\Lx=\mathbb{E}[\|\hat{x}_\theta-x\|_2^2]$) to examine loss weighting with the same direct output.

Figure~\ref{fig:loss-weighting} compares the convergence trajectories alongside the direct velocity baseline. At 100K steps, clean output under unweighted $\Lx$ reaches \FID{} 3.48 and \IS{} 204.9. At 250K steps, it reaches \FID{} 2.38 and \IS{} 256.88, compared with 2.84 and 204.83 for clean output under $\Lv$. At B/1, clean prediction performs strongly under both velocity-weighted and unweighted clean regression, with the latter reaching \FID{} 2.38.

\subsection{Reference Models}
\label{sec:reference}

\begin{table}[t]
    \centering
    \footnotesize
    \setlength{\tabcolsep}{4.5pt}
    \begin{tabular}{lccccc}
    \toprule
    Model & Space & Extra rep. & Params & \FID{} $\downarrow$ & \IS{} $\uparrow$ \\
    \midrule
    DiT-XL/2~\citep{peebles2023dit} & VAE & -- & 675M & 2.27 & 278.20 \\
    SiT-XL/2~\citep{ma2024sit} & VAE & -- & 675M & 2.06 & 277.50 \\
    REPA-SiT-XL/2~\citep{yu2025repa} & VAE & DINOv2 & 675M & 1.42 & 305.70 \\
    MDTv2-XL/2~\citep{gao2024mdtv2} & VAE & -- & 675.8M & 1.58 & 314.73 \\
    DiffiT~\citep{hatamizadeh2023diffit} & VAE & -- & 561M & 1.73 & 276.49 \\
    JiT-H/16~\citep{li2025jit} & pixel & -- & 953M & 1.86 & 303.40 \\
    RiT~\citep{zhang2026rit} & DINOv2 & DINOv2 & 676M & \textbf{1.14} & -- \\
    \midrule
    \method{}-H/1 ($\Lv$) & VAE & -- & 951.3M & 1.19 & 271.96 \\
    \bottomrule
    \end{tabular}
    \vspace{5pt}
    \caption{\textbf{Representative guided ImageNet $256\times256$ models.} Baselines from cited reports; ``Extra rep.'' denotes external self-supervised representations.}
    \label{tab:comparison}
\end{table}

Table~\ref{tab:comparison} contextualizes \method{}-H/1 alongside representative guided ImageNet $256\times256$ models. At Huge scale, \method{}-H/1 reaches \FID{} 1.19 and \IS{} 271.96. Figure~\ref{fig:samples} (Appendix~\ref{app:samples}) provides paired qualitative comparisons on B/1.

\vspace{5pt}

\FloatBarrier
\section{Discussion}

\vspace{5pt}

\paragraph{Parameterization as a computation boundary.}
The unregularized clean-to-velocity identity is exact on $0<t<1$, but its placement across the model boundary is consequential. Direct velocity embeds the vector-field response inside the Transformer, whereas clean prediction moves the known input residual and gain outside it. For a fixed map $T$, learning $T\!\circ\!f_\theta$ versus unconstrained $g_\theta$ yields distinct inductive biases: algebraic equivalence dictates what can be represented, while parameterization dictates what is learned.

\paragraph{Preserving structure in learned representations.}
Autoencoders compress pixels without yielding isotropic latents. FLUX latents exhibit strong spectral concentration, which clean prediction presents directly to the network; velocity mixes this signal with a uniform noise floor. Beyond altering autoencoders, the prediction target determines whether intrinsic representation geometry is preserved during flow training.

\paragraph{Finite learning versus solver semantics.}
Both parameterizations expose velocity to the same solver. The FID margin narrows from Base to Large, consistent with the finite-capacity view that externalizing the analytic readout is most beneficial when model capacity is limited. The Huge result shows that the clean-output model remains effective at larger scale.

\vspace{5pt}

\section{Conclusion}

\vspace{5pt}

A flow solver requires velocity values, but a generative Transformer need not learn velocity directly. Predicting the clean latent with an exact affine readout preserves target geometry, improves ImageNet generation across Base and Large scales under matched velocity objectives, and scales to 951M parameters, reaching an \FID{} of 1.19 at Huge scale. An ablation at Base scale shows that direct clean regression ($\Lx$) remains highly effective, achieving \FID{} 2.38 without time-dependent error weighting. Algebraically equivalent targets yield unequal learning problems when known linear computation is externalized from the network.

\clearpage

\bibliography{paper}

@inproceedings{ho2020denoising,
  title = {Denoising Diffusion Probabilistic Models},
  author = {Ho, Jonathan and Jain, Ajay and Abbeel, Pieter},
  booktitle = {NeurIPS},
  year = {2020},
  url = {https://arxiv.org/abs/2006.11239}
}

@inproceedings{nichol2021improved,
  title = {Improved Denoising Diffusion Probabilistic Models},
  author = {Nichol, Alex and Dhariwal, Prafulla},
  booktitle = {ICML},
  year = {2021},
  url = {https://arxiv.org/abs/2102.09672}
}

@inproceedings{salimans2022progressive,
  title = {Progressive Distillation for Fast Sampling of Diffusion Models},
  author = {Salimans, Tim and Ho, Jonathan},
  booktitle = {ICLR},
  year = {2022},
  url = {https://arxiv.org/abs/2202.00512}
}

@article{ho2022cfg,
  title = {Classifier-Free Diffusion Guidance},
  author = {Ho, Jonathan and Salimans, Tim},
  journal = {arXiv preprint arXiv:2207.12598},
  year = {2022},
  url = {https://arxiv.org/abs/2207.12598}
}

@inproceedings{karras2022elucidating,
  title = {Elucidating the Design Space of Diffusion-Based Generative Models},
  author = {Karras, Tero and Aittala, Miika and Aila, Timo and Laine, Samuli},
  booktitle = {NeurIPS},
  year = {2022},
  url = {https://arxiv.org/abs/2206.00364}
}

@inproceedings{choi2022perception,
  title = {Perception Prioritized Training of Diffusion Models},
  author = {Choi, Jooyoung and Lee, Jungbeom and Shin, Chaehun and Kim, Sungwon and Kim, Hyunwoo J. and Yoon, Sungroh},
  booktitle = {CVPR},
  year = {2022},
  url = {https://arxiv.org/abs/2204.00227}
}

@inproceedings{hang2023efficient,
  title = {Efficient Diffusion Training via {Min-SNR} Weighting Strategy},
  author = {Hang, Tiankai and Gu, Shuyang and Li, Chen and Bao, Jianmin and Chen, Dong and Hu, Han and Geng, Xin and Guo, Baining},
  booktitle = {ICCV},
  year = {2023},
  url = {https://arxiv.org/abs/2303.09556}
}

@inproceedings{kingma2023understanding,
  title = {Understanding Diffusion Objectives as the {ELBO} with Simple Data Augmentation},
  author = {Kingma, Diederik P. and Gao, Ruiqi},
  booktitle = {NeurIPS},
  year = {2023},
  url = {https://arxiv.org/abs/2303.00848}
}

@inproceedings{li2025jit,
  title = {Back to Basics: Let Denoising Generative Models Denoise},
  author = {Li, Tianhong and He, Kaiming},
  booktitle = {CVPR},
  year = {2026},
  url = {https://arxiv.org/abs/2511.13720}
}

@inproceedings{jin2026revisiting,
  title = {Revisiting Diffusion Model Predictions Through Dimensionality},
  author = {Jin, Qing and Wang, Chaoyang},
  booktitle = {ICML},
  year = {2026},
  url = {https://arxiv.org/abs/2601.21419}
}

@article{gagneux2026training,
  title = {Training Flow Matching: The Role of Weighting and Parameterization},
  author = {Gagneux, Anne and Martin, S{\'e}gol{\`e}ne and Gribonval, R{\'e}mi and Massias, Mathurin},
  journal = {arXiv preprint arXiv:2603.06454},
  year = {2026},
  url = {https://arxiv.org/abs/2603.06454}
}

@inproceedings{lipman2023flow,
  title = {Flow Matching for Generative Modeling},
  author = {Lipman, Yaron and Chen, Ricky T. Q. and Ben-Hamu, Heli and Nickel, Maximilian and Le, Matt},
  booktitle = {ICLR},
  year = {2023},
  url = {https://arxiv.org/abs/2210.02747}
}

@inproceedings{liu2023flow,
  title = {Flow Straight and Fast: Learning to Generate and Transfer Data with Rectified Flow},
  author = {Liu, Xingchao and Gong, Chengyue and Liu, Qiang},
  booktitle = {ICLR},
  year = {2023},
  url = {https://arxiv.org/abs/2209.03003}
}

@inproceedings{albergo2023building,
  title = {Building Normalizing Flows with Stochastic Interpolants},
  author = {Albergo, Michael S. and Vanden-Eijnden, Eric},
  booktitle = {ICLR},
  year = {2023},
  url = {https://arxiv.org/abs/2209.15571}
}

@article{albergo2023stochastic,
  title = {Stochastic Interpolants: A Unifying Framework for Flows and Diffusions},
  author = {Albergo, Michael S. and Boffi, Nicholas M. and Vanden-Eijnden, Eric},
  journal = {arXiv preprint arXiv:2303.08797},
  year = {2023},
  url = {https://arxiv.org/abs/2303.08797}
}

@inproceedings{ma2024sit,
  title = {SiT: Exploring Flow and Diffusion-Based Generative Models with Scalable Interpolant Transformers},
  author = {Ma, Nanye and Goldstein, Mark and Albergo, Michael S. and Boffi, Nicholas M. and Vanden-Eijnden, Eric and Xie, Saining},
  booktitle = {ECCV},
  year = {2024},
  url = {https://arxiv.org/abs/2401.08740}
}

@inproceedings{esser2024scaling,
  title = {Scaling Rectified Flow Transformers for High-Resolution Image Synthesis},
  author = {Esser, Patrick and Kulal, Sumith and Blattmann, Andreas and Entezari, Rahim and M{\"u}ller, Jonas and Saini, Harry and Levi, Yam and Lorenz, Dominik and Sauer, Axel and Boesel, Frederic and Podell, Dustin and Dockhorn, Tim and English, Zion and Lacey, Kyle and Goodwin, Alex and Marek, Yannik and Rombach, Robin},
  booktitle = {ICML},
  year = {2024},
  url = {https://arxiv.org/abs/2403.03206}
}

@inproceedings{rombach2022high,
  title = {High-Resolution Image Synthesis with Latent Diffusion Models},
  author = {Rombach, Robin and Blattmann, Andreas and Lorenz, Dominik and Esser, Patrick and Ommer, Bj{\"o}rn},
  booktitle = {CVPR},
  year = {2022},
  url = {https://arxiv.org/abs/2112.10752}
}

@inproceedings{bao2023all,
  title = {All are Worth Words: A ViT Backbone for Diffusion Models},
  author = {Bao, Fan and Nie, Shen and Xue, Kaiwen and Cao, Yue and Li, Chongxuan and Su, Hang and Zhu, Jun},
  booktitle = {CVPR},
  year = {2023},
  url = {https://arxiv.org/abs/2209.12152}
}

@inproceedings{peebles2023dit,
  title = {Scalable Diffusion Models with Transformers},
  author = {Peebles, William and Xie, Saining},
  booktitle = {ICCV},
  year = {2023},
  url = {https://arxiv.org/abs/2212.09748}
}

@article{gao2024mdtv2,
  title = {{MDTv2}: Masked Diffusion Transformer is a Strong Image Synthesizer},
  author = {Gao, Shanghua and Zhou, Pan and Cheng, Ming-Ming and Yan, Shuicheng},
  journal = {arXiv preprint arXiv:2303.14389v2},
  year = {2024},
  url = {https://arxiv.org/abs/2303.14389v2}
}

@inproceedings{hatamizadeh2023diffit,
  title = {DiffiT: Diffusion Vision Transformers for Image Generation},
  author = {Hatamizadeh, Ali and Song, Jiaming and Liu, Guilin and Kautz, Jan and Vahdat, Arash},
  booktitle = {ECCV},
  year = {2024},
  url = {https://arxiv.org/abs/2312.02139}
}

@inproceedings{chen2023pixart,
  title = {PixArt-Alpha: Fast Training of Diffusion Transformer for Photorealistic Text-to-Image Synthesis},
  author = {Chen, Junsong and Yu, Jincheng and Ge, Chongjian and Yao, Lewei and Xie, Enze and Wu, Yue and Wang, Zhongdao and Kwok, James and Luo, Ping and Lu, Huchuan and Li, Zhenguo},
  booktitle = {ICLR},
  year = {2024},
  url = {https://arxiv.org/abs/2310.00426}
}

@article{oquab2024dinov2,
  title = {DINOv2: Learning Robust Visual Features without Supervision},
  author = {Oquab, Maxime and Darcet, Timoth{\'e}e and Moutakanni, Th{\'e}o and Vo, Huy and Szafraniec, Marc and Khalidov, Vasil and Fernandez, Pierre and Haziza, Daniel and Massa, Francisco and El-Nouby, Alaaeldin and Assran, Mahmoud and Ballas, Nicolas and Galuba, Wojciech and Howes, Russell and Huang, Po-Yao and Li, Shang-Wen and Misra, Ishan and Rabbat, Michael and Sharma, Vasu and Synnaeve, Gabriel and Xu, Hu and J{\'e}gou, Herv{\'e} and Mairal, Julien and Labatut, Patrick and Joulin, Armand and Bojanowski, Piotr},
  journal = {Transactions on Machine Learning Research},
  year = {2024},
  url = {https://arxiv.org/abs/2304.07193}
}

@inproceedings{yu2025repa,
  title = {Representation Alignment for Generation: Training Diffusion Transformers Is Easier Than You Think},
  author = {Yu, Sihyun and Kwak, Sangkyung and Jang, Huiwon and Jeong, Jongheon and Huang, Jonathan and Shin, Jinwoo and Xie, Saining},
  booktitle = {ICLR},
  year = {2025},
  url = {https://arxiv.org/abs/2410.06940}
}

@inproceedings{zheng2025diffusion,
  title = {Diffusion Transformers with Representation Autoencoders},
  author = {Zheng, Boyang and Ma, Nanye and Tong, Shengbang and Xie, Saining},
  booktitle = {ICLR},
  year = {2026},
  url = {https://arxiv.org/abs/2510.11690}
}

@article{singh2026improved,
  title = {Improved Baselines with Representation Autoencoders},
  author = {Singh, Jaskirat and Zheng, Boyang and Wu, Zongze and Zhang, Richard and Shechtman, Eli and Xie, Saining},
  journal = {arXiv preprint arXiv:2605.18324},
  year = {2026},
  url = {https://arxiv.org/abs/2605.18324}
}

@article{zhang2026rit,
  title = {RiT: Vanilla Diffusion Transformers Suffice in Representation Space},
  author = {Zhang, Le and Mang, Ning and Agrawal, Aishwarya},
  journal = {arXiv preprint arXiv:2605.21981},
  year = {2026},
  url = {https://arxiv.org/abs/2605.21981}
}

@inproceedings{wang2026vaerepa,
  title = {{SRA 2}: Variational Autoencoder Self-Representation Alignment for Efficient Diffusion Training},
  author = {Wang, Mengmeng and Jiang, Dengyang and Li, Liuzhuozheng and Lin, Yucheng and Shen, Guojiang and Kong, Xiangjie and Liu, Yong and Dai, Guang and Wang, Jingdong},
  booktitle = {CVPR},
  year = {2026},
  url = {https://arxiv.org/abs/2601.17830}
}

@inproceedings{vincent2008extracting,
  title = {Extracting and Composing Robust Features with Denoising Autoencoders},
  author = {Vincent, Pascal and Larochelle, Hugo and Bengio, Yoshua and Manzagol, Pierre-Antoine},
  booktitle = {ICML},
  year = {2008},
  url = {https://dl.acm.org/doi/10.1145/1390156.1390294}
}

@article{vincent2010stacked,
  title = {Stacked Denoising Autoencoders: Learning Useful Representations in a Deep Network with a Local Denoising Criterion},
  author = {Vincent, Pascal and Larochelle, Hugo and Lajoie, Isabelle and Bengio, Yoshua and Manzagol, Pierre-Antoine},
  journal = {Journal of Machine Learning Research},
  year = {2010},
  url = {https://www.jmlr.org/papers/v11/vincent10a.html}
}

@inproceedings{bengio2013generalized,
  title = {Generalized Denoising Auto-Encoders as Generative Models},
  author = {Bengio, Yoshua and Yao, Li and Alain, Guillaume and Vincent, Pascal},
  booktitle = {NeurIPS},
  year = {2013},
  url = {https://arxiv.org/abs/1305.6663}
}

@article{alain2014regularized,
  title = {What Regularized Auto-Encoders Learn from the Data-Generating Distribution},
  author = {Alain, Guillaume and Bengio, Yoshua},
  journal = {Journal of Machine Learning Research},
  year = {2014},
  url = {https://www.jmlr.org/papers/v15/alain14a.html}
}

@article{zhang2017beyond,
  title = {Beyond a Gaussian Denoiser: Residual Learning of Deep CNN for Image Denoising},
  author = {Zhang, Kai and Zuo, Wangmeng and Chen, Yunjin and Meng, Deyu and Zhang, Lei},
  journal = {IEEE Transactions on Image Processing},
  year = {2017},
  url = {https://arxiv.org/abs/1608.03981}
}

@inproceedings{pope2021intrinsic,
  title = {The Intrinsic Dimension of Images and Its Impact on Learning},
  author = {Pope, Phillip and Zhu, Chen and Abdelkader, Ahmed and Goldblum, Micah and Goldstein, Tom},
  booktitle = {ICLR},
  year = {2021},
  url = {https://arxiv.org/abs/2104.08894}
}

@inproceedings{stanczuk2024diffusion,
  title = {Diffusion Models Encode the Intrinsic Dimension of Data Manifolds},
  author = {Stanczuk, Jan Pawel and Batzolis, Georgios and Deveney, Teo and Sch{\"o}nlieb, Carola-Bibiane},
  booktitle = {ICML},
  year = {2024},
  url = {https://arxiv.org/abs/2212.12611}
}

@article{lu2022dpmsolverpp,
  title = {{DPM-Solver++}: Fast Solver for Guided Sampling of Diffusion Probabilistic Models},
  author = {Lu, Cheng and Zhou, Yuhao and Bao, Fan and Chen, Jianfei and Li, Chongxuan and Zhu, Jun},
  journal = {Machine Intelligence Research},
  year = {2025},
  url = {https://doi.org/10.1007/s11633-025-1562-4}
}

@inproceedings{heusel2017fid,
  title = {GANs Trained by a Two Time-Scale Update Rule Converge to a Local Nash Equilibrium},
  author = {Heusel, Martin and Ramsauer, Hubert and Unterthiner, Thomas and Nessler, Bernhard and Hochreiter, Sepp},
  booktitle = {NeurIPS},
  year = {2017},
  url = {https://arxiv.org/abs/1706.08500}
}

@inproceedings{salimans2016is,
  title = {Improved Techniques for Training GANs},
  author = {Salimans, Tim and Goodfellow, Ian and Zaremba, Wojciech and Cheung, Vicki and Radford, Alec and Chen, Xi},
  booktitle = {NeurIPS},
  year = {2016},
  url = {https://arxiv.org/abs/1606.03498}
}

@misc{blackforestlabs2026flux2small,
  title = {{FLUX.2 Small Decoder}},
  author = {{Black Forest Labs}},
  year = {2026},
  howpublished = {\href{https://huggingface.co/black-forest-labs/FLUX.2-small-decoder}{Hugging Face model repository}}
}

@inproceedings{deng2009imagenet,
  title = {{ImageNet}: A Large-Scale Hierarchical Image Database},
  author = {Deng, Jia and Dong, Wei and Socher, Richard and Li, Li-Jia and Li, Kai and Fei-Fei, Li},
  booktitle = {CVPR},
  year = {2009}
}

@inproceedings{loshchilov2019adamw,
  title = {Decoupled Weight Decay Regularization},
  author = {Loshchilov, Ilya and Hutter, Frank},
  booktitle = {ICLR},
  year = {2019}
}

@article{russakovsky2015imagenet,
  title = {{ImageNet} Large Scale Visual Recognition Challenge},
  author = {Russakovsky, Olga and Deng, Jia and Su, Hao and Krause, Jonathan and Satheesh, Sanjeev and Ma, Sean and Huang, Zhiheng and Karpathy, Andrej and Khosla, Aditya and Bernstein, Michael and Berg, Alexander C. and Fei-Fei, Li},
  journal = {International Journal of Computer Vision},
  year = {2015}
}

@article{wu2025qwenimagetechnicalreport,
  title = {{Qwen-Image} Technical Report},
  author = {Wu, Chenfei and Li, Jiahao and Zhou, Jingren and Lin, Junyang and Gao, Kaiyuan and Yan, Kun and Yin, Sheng-ming and Bai, Shuai and Xu, Xiao and Chen, Yilei and Chen, Yuxiang and Tang, Zecheng and Zhang, Zekai and Wang, Zhengyi and Yang, An and Yu, Bowen and Cheng, Chen and Liu, Dayiheng and Li, Deqing and Zhang, Hang and Meng, Hao and Wei, Hu and Ni, Jingyuan and Chen, Kai and Cao, Kuan and Peng, Liang and Qu, Lin and Wu, Minggang and Wang, Peng and Yu, Shuting and Wen, Tingkun and Feng, Wensen and Xu, Xiaoxiao and Wang, Yi and Zhang, Yichang and Zhu, Yongqiang and Wu, Yujia and Cai, Yuxuan and Liu, Zenan},
  journal = {arXiv preprint arXiv:2508.02324},
  year = {2025},
  url = {https://arxiv.org/abs/2508.02324}
}
\bibliographystyle{iclr2027_conference}

\clearpage
\appendix
\renewcommand{\theHsection}{supp.\Alph{section}}
\renewcommand{\theHsubsection}{supp.\Alph{section}.\arabic{subsection}}

\makeatletter
\setlength{\@fptop}{0pt}
\setlength{\@fpsep}{14pt}
\setlength{\@fpbot}{0pt plus 1fil}
\makeatother

\thispagestyle{plain}
{\noindent\fontsize{24}{29}\selectfont\textbf{Appendix}\par}
\vspace{10pt}
{\normalsize\noindent This appendix collects target conversions, theoretical derivations, implementation details, cross-tokenizer generalization analyses, classifier-free guidance sweeps, and uncurated visual samples that support the main paper.\par}
\vspace{5pt}
\begingroup
\hypersetup{linkcolor=black}
\etoctoccontentsline{part}{}
\etocstandardlines
\etocsettocstyle{}{}
\etocsetnexttocdepth{subsection}
\localtableofcontents
\endgroup

\clearpage

\section{Target Conversions}
\label{app:conversions}

For fixed $0<t<1$, any one of $x$, $\epsilon$, and $v=x-\epsilon$ determines the other two from $z_t$. For clean prediction,
\begin{align}
    \hat{\epsilon}^{(x)}_\theta&=\frac{z_t-t\hat{x}_\theta}{1-t}, &
    \hat{v}^{(x)}_\theta&=\frac{\hat{x}_\theta-z_t}{1-t}.
\end{align}
For noise prediction,
\begin{align}
    \hat{x}^{(\epsilon)}_\theta&=\frac{z_t-(1-t)\hat{\epsilon}_\theta}{t}, &
    \hat{v}^{(\epsilon)}_\theta&=\frac{z_t-\hat{\epsilon}_\theta}{t}.
\end{align}
For velocity prediction,
\begin{align}
    \hat{x}^{(v)}_\theta&=z_t+(1-t)\hat{v}_\theta, &
    \hat{\epsilon}^{(v)}_\theta&=z_t-t\hat{v}_\theta.
\end{align}
These identities also expose error scaling. Let $e_x=\hat{x}_\theta-x$, $e_\epsilon=\hat{\epsilon}_\theta-\epsilon$, and $e_v=\hat{v}_\theta-v$. Then
\begin{align}
    \hat{v}^{(x)}_\theta-v&=\frac{e_x}{1-t}, &
    \hat{\epsilon}^{(x)}_\theta-\epsilon&=-\frac{t}{1-t}e_x,\\
    \hat{x}^{(\epsilon)}_\theta-x&=-\frac{1-t}{t}e_\epsilon, &
    \hat{v}^{(\epsilon)}_\theta-v&=-\frac{1}{t}e_\epsilon,\\
    \hat{x}^{(v)}_\theta-x&=(1-t)e_v, &
    \hat{\epsilon}^{(v)}_\theta-\epsilon&=-t e_v.
\end{align}
The targets are convertible, but equal MSE in one target is not equal MSE in another.

\clearpage

\section{Gaussian Derivation}
\label{app:proof}

Let $x_i\sim\mathcal{N}(0,\lambda_i)$ and $\epsilon_i\sim\mathcal{N}(0,1)$ be independent in an eigenbasis of $\Sigma$. With $z_i=t x_i+(1-t)\epsilon_i$ and $q_i=t^2\lambda_i+(1-t)^2$, joint Gaussian conditioning gives
\begin{equation}
    \operatorname{Var}(a\mid z)
    =\operatorname{Var}(a)-\frac{\operatorname{Cov}(a,z)^2}{\operatorname{Var}(z)}.
\end{equation}
Using $\operatorname{Cov}(x_i,z_i)=t\lambda_i$, $\operatorname{Cov}(\epsilon_i,z_i)=1-t$, and $\operatorname{Cov}(v_i,z_i)=t\lambda_i-(1-t)$ yields
\begin{align}
    \operatorname{Var}(x_i\mid z_i)&=\frac{\lambda_i(1-t)^2}{q_i},\\
    \operatorname{Var}(\epsilon_i\mid z_i)&=\frac{t^2\lambda_i}{q_i},\\
    \operatorname{Var}(v_i\mid z_i)&=\frac{\lambda_i}{q_i}.
\end{align}
Therefore $\operatorname{Var}(v_i\mid z_i)=\operatorname{Var}(x_i\mid z_i)/(1-t)^2$. Summing across directions gives local Bayes residual risks
\begin{align}
    \mathcal{R}_{x}(t)&=\sum_i\frac{\lambda_i(1-t)^2}{t^2\lambda_i+(1-t)^2},\\
    \mathcal{R}_{v}(t)&=\sum_i\frac{\lambda_i}{t^2\lambda_i+(1-t)^2}.
\end{align}
Together with Proposition~\ref{prop:geometry}, these expressions quantify how the same sample-wise target conversion changes conditional scale and regression gain under the local Gaussian model.

\paragraph{Spectral-mixture identity.}
Adding $I$ preserves the covariance eigenvectors and shifts each eigenvalue by one. The leading-$k$ velocity variance is therefore $\sum_{i=1}^{k}\lambda_i+k$, and total velocity variance is $S+D$. Dividing gives Eq.~\eqref{eq:spectral-mixture}. For descending eigenvalues, the mean of the first $k$ values is at least the mean of all $D$ values, so $C_x(k)\geq k/D$. Substitution yields $C_v(k)\leq C_x(k)$, with equality for an isotropic spectrum.

\clearpage

\section{Implementation Details}
\label{app:implementation}

\paragraph{Representation.}
Images are encoded deterministically with the frozen FLUX.2 VAE. The implementation patchifies the VAE output into a 128-channel, $16\times16$ latent grid and normalizes each channel using frozen running statistics. The decoder is used only after ODE integration.

\paragraph{Architecture.}
Table~\ref{tab:architecture} gives the model family. Each block uses self-attention, a swish-gated linear unit (SwiGLU) feed-forward network, root mean square (RMS) normalization, rotary position embeddings, and adaptive time/class modulation. The output projection reconstructs the latent grid. Attention and projection dropout rates are zero.

\begin{table}[htbp]
\centering
\small
\begin{tabular}{lrrrr}
\toprule
Model & Depth & Width & Heads & Params \\
\midrule
\method{}-B/1 & 12 & 768 & 12 & 130.5M \\
\method{}-L/1 & 24 & 1024 & 16 & 458.1M \\
\method{}-H/1 & 32 & 1280 & 16 & 951.3M \\
\bottomrule
\end{tabular}
\caption{\textbf{\method{} architecture family.} Parameter counts include the trainable generator and exclude the frozen VAE.}
\label{tab:architecture}
\end{table}

\paragraph{Optimization.}
We use AdamW~\citep{loshchilov2019adamw} with $\beta_1=0.9$, $\beta_2=0.95$, zero weight decay, base learning rate $5\times10^{-5}$, and effective learning rate $2\times10^{-4}$ after batch scaling. Training lasts 200 epochs with five warmup epochs. Exponential moving averages (EMA) use decay 0.9999 and 0.9996. The class condition is independently dropped with probability 0.1.

\paragraph{Time and sampling.}
The exact conversion identities above concern the unregularized field on $0<t<1$. To distinguish endpoint conventions, define $d_\epsilon(t)=\max(1-t,t_\epsilon)$. If both the clean-output prediction and its training target use this denominator, their residual is
\begin{equation}
    \frac{\hat x-z_t}{d_\epsilon(t)}-\frac{x-z_t}{d_\epsilon(t)}
    =\frac{\hat x-x}{d_\epsilon(t)}.
\end{equation}
The resulting clean-error weight is $d_\epsilon(t)^{-2}$. A sampling readout with the same denominator equals the unregularized field multiplied by $(1-t)/d_\epsilon(t)$, with exact agreement for $t\leq1-t_\epsilon$.
Training samples $\operatorname{logit}(t)\sim\mathcal{N}(-0.8,0.8^2)$. Division by $1-t$ is clipped at $t_\epsilon=0.05$. Evaluation uses 50 Heun steps, 50K generated images, classifier-free guidance~\citep{ho2022cfg}, and guidance interval $[0.1,1]$. Training trajectories and qualitative samples use the same sampling settings within each comparison. Appendix~\ref{app:cfg-sweep} reports guidance sweeps.

\clearpage

\section{Cross-Tokenizer Generalization with Qwen-VAE}
\label{app:qwen-vae}

We repeat the matched B/1 comparison on the pretrained Qwen-VAE representation~\citep{wu2025qwenimagetechnicalreport} to examine transfer across autoencoder representations. Both models use 200 epochs (250K steps), velocity MSE ($\Lv$), the same logit-normal time distribution, and the same sampling settings.

\begin{figure}[!htbp]
    \centering
    \includegraphics[width=0.62\linewidth]{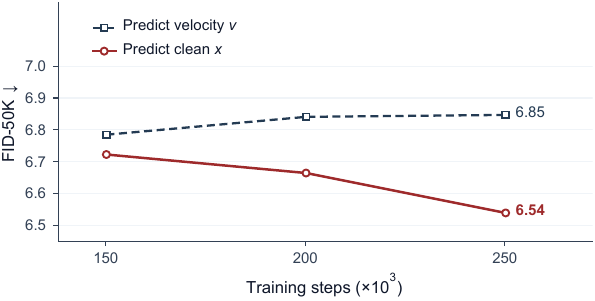}
    \caption{\textbf{Matched training dynamics on Qwen-VAE.} FID-50K of matched B/1 models on Qwen-VAE at 150K, 200K, and 250K steps. Clean-$x$ prediction has lower FID at all three measured checkpoints, ending at 6.54 versus 6.85.}
    \label{fig:qwen-training}
\end{figure}

Figure~\ref{fig:qwen-training} and Table~\ref{tab:qwen-vae} show the convergence behavior and final metrics. At the measured checkpoints from 150K to 250K steps, clean-$x$ prediction has lower \FID{} than velocity prediction. The margin grows from 6.72 vs.\ 6.79 at 150K steps to 6.54 vs.\ 6.85 at 250K steps (Table~\ref{tab:qwen-vae}). The same FID ordering is observed on Qwen-VAE.

\begin{table}[!htbp]
\centering
\small
\begin{tabular}{llccc}
\toprule
Model & Representation & Direct Output & Steps & \FID{} $\downarrow$ \\
\midrule
\method{}-B/1 & Qwen-VAE & clean latent $x$ & 150K & 6.72 \\
\method{}-B/1 & Qwen-VAE & velocity $v$     & 150K & 6.79 \\
\midrule
\method{}-B/1 & Qwen-VAE & clean latent $x$ & 200K & 6.67 \\
\method{}-B/1 & Qwen-VAE & velocity $v$     & 200K & 6.84 \\
\midrule
\method{}-B/1 & Qwen-VAE & clean latent $x$ & 250K & \textbf{6.54} \\
\method{}-B/1 & Qwen-VAE & velocity $v$     & 250K & 6.85 \\
\bottomrule
\end{tabular}
\caption{\textbf{Matched parameterization comparison on Qwen-VAE.} Both models are B/1 configurations trained under the same velocity MSE objective ($\Lv$) and evaluated under the same sampling settings across convergence checkpoints. Clean prediction achieves lower FID at every evaluated stage, ending at 6.54 vs.\ 6.85 at 250K steps.}
\label{tab:qwen-vae}
\end{table}

\clearpage

\section{Classifier-Free Guidance Sweep}
\label{app:cfg-sweep}

\begin{figure}[!htbp]
\centering
\includegraphics[width=\linewidth]{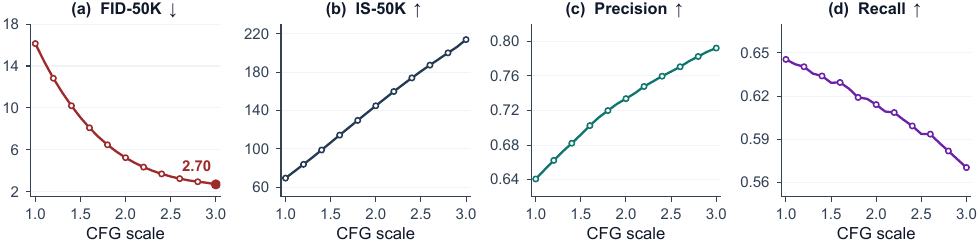}
\caption{\textbf{Classifier-free guidance sweep for \method{}-B/1.} FID, IS, precision, and recall across the evaluated CFG grid.}
\label{fig:b1-cfg-sweep}
\end{figure}

\paragraph{\method{}-B/1 guidance sweep.}
We evaluate the 200-epoch \method{}-B/1 clean-output model across CFG scales from 1.0 to 3.0 in increments of 0.1. The lowest observed \FID{} is 2.70 at CFG 3.0, with \IS{} 213.89. Figure~\ref{fig:b1-cfg-sweep} shows the full sweep.

\begin{figure}[!htbp]
\centering
\includegraphics[width=\linewidth]{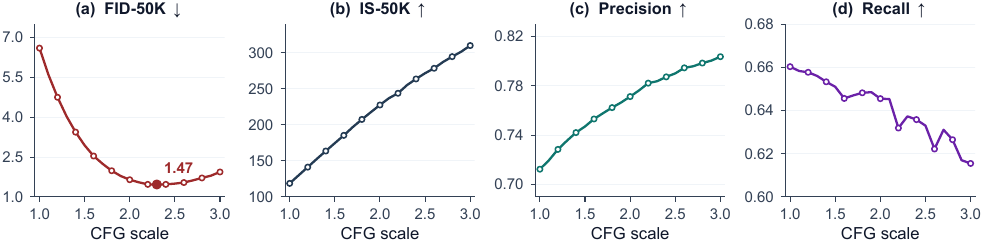}
\caption{\textbf{Classifier-free guidance sweep for \method{}-L/1.} FID, IS, precision, and recall across the evaluated CFG grid; clean-$x$ achieves an lowest observed \FID{} of 1.47 at CFG 2.3.}
\label{fig:l1-cfg-sweep}
\end{figure}

\paragraph{\method{}-L/1 guidance sweep.}
We evaluate the 200-epoch \method{}-L/1 clean-output model across CFG scales from 1.0 to 3.0 in increments of 0.1. The lowest observed \FID{} is 1.47 at CFG 2.3, with \IS{} 255.28. Figure~\ref{fig:l1-cfg-sweep} shows the full sweep.

\begin{figure}[!htbp]
    \centering
    \includegraphics[width=\linewidth]{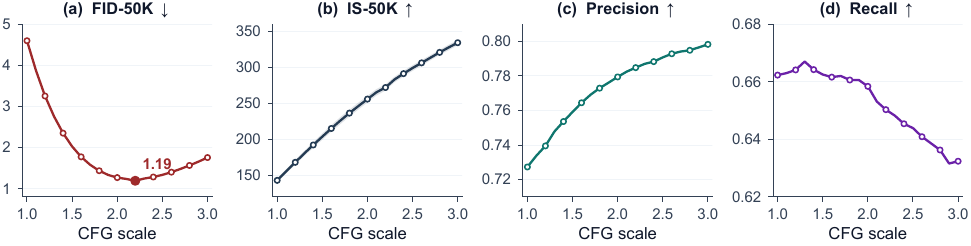}
    \caption{\textbf{Classifier-free guidance sweep for \method{}-H/1.} FID, IS, precision, and recall across the evaluated CFG grid; the IS band shows one standard deviation.}
    \label{fig:cfg-sweep}
\end{figure}

\paragraph{\method{}-H/1 guidance sweep.}
We evaluate the 200-epoch \method{}-H/1 clean-output model across CFG scales from 1.0 to 3.0 in increments of 0.1. The lowest observed \FID{} is 1.193 at CFG 2.2, with \IS{} 271.96. Figure~\ref{fig:cfg-sweep} shows the full sweep.

\clearpage

\section{Qualitative Visual Samples and Comparison}
\label{app:samples}

\begin{figure}[!htbp]
    \centering
    \includegraphics[width=\linewidth]{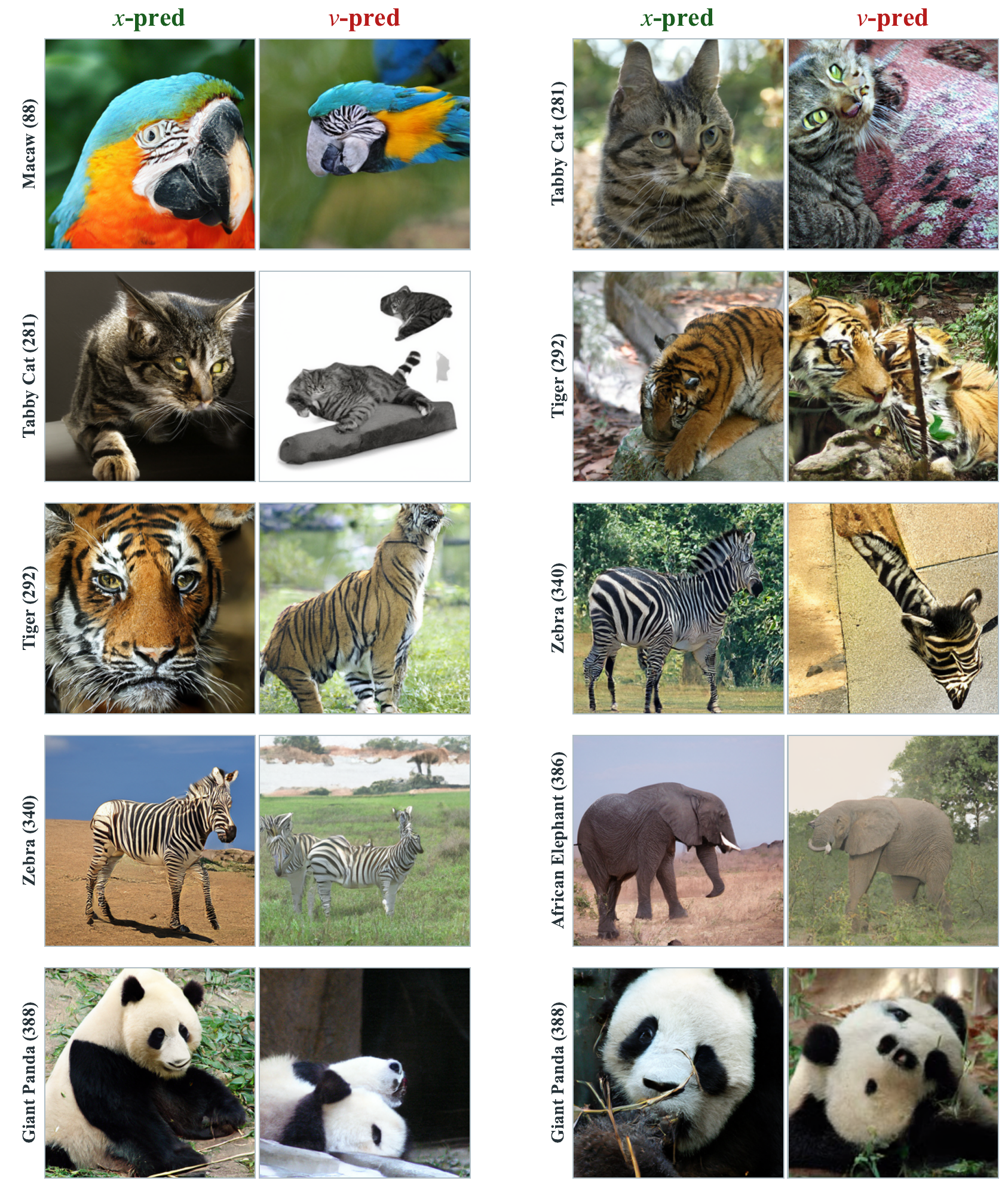}
    \caption{\textbf{Additional paired qualitative comparison on ImageNet $256\times256$ under matched B/1 scale.} Generations from matched \method{}-B/1 models trained for 200 epochs and sampled under the same settings. Left sub-columns ($x$-pred, green) show clean prediction; right sub-columns ($v$-pred, red) show direct velocity prediction across 10 paired class-conditional samples.}
    \label{fig:samples}
\end{figure}

Figure~\ref{fig:samples} displays paired class-conditional ImageNet $256\times256$ generations produced by matched \method{}-B/1 models with clean-$x$ and direct velocity-$v$ prediction under the same sampling settings. The paired samples illustrate differences in structure and sharpness across the displayed animal classes.

\end{document}